# The IQ of Artificial Intelligence

Dimiter Dobrev
Institute of Mathematics and Informatics
Bulgarian Academy of Sciences
*d@dobrev.com*

All it takes to identify the computer programs which are Artificial Intelligence is to give them a test and award AI to those that pass the test. Let us say that the scores they earn at the test will be called IQ. We cannot pinpoint a minimum IQ threshold that a program has to cover in order to be AI, however, we will choose a certain value. Thus, our definition for AI will be any program the IQ of which is above the chosen value. While this idea has already been implemented in [3], here we will revisit this construct in order to introduce certain improvements.

## Introduction

We will use a test to determine what AI is. The test will produce a certain score, and we say that this is the program's IQ. Then we decide that all computer programs the IQ of which is above a certain level satisfy the AI definition.

In order to explain this concept, let us make an analogy with the admission tests given to candidates who wish to become university students. The problems given at the test are selected randomly, but all candidate students receive the same problems. Withal, solving the problems should require logical thinking, because we aim to enroll students who think logically rather than the lucky ones that may hit the right answers haphazardly. The score is based on the number of problems solved by each candidate student. We cannot say how many problems should be solved, because we do not know how many candidates will show up at the test, nor do we know how well or unwell they will perform. We may say set a certain score (e.g. 4.50) and say that we intend to admit all candidates who earn more than 4.50. However, it would be better to fix the minimum score after the test is done. Then we will take the score earned by the n-th ranking candidate (e.g. if n is 100 we take the score of the candidate whose score puts him in the 100$^{th}$ position in the ranking and thus we enroll the top 100 candidates).

This analogy describes well the AI test, but when the candidates are computer programs, we cannot select the 100 top performers, because in this case there will be infinitely many candidates. A better analogy is perhaps a recruitment contest for the CEO of a corporation with a test that drags on over time. The test will stop when a candidate earns a sufficient number of points. How many points are enough? While we may select a certain level in advance, we can adjust this level later if the initial level turns out to be too low or too high.

A similar construct was already put forward in [3] whereby the various programs were given different scores (IQ∈[0.1]). Here we will revisit this construct in order to improve it. The reasons for which [3] needs improvement are:

1. In [3] we dealt concurrently with the questions of 'What is AI?' and 'How can we create AI?'. Mixing up 'what' and 'how' is not a good idea. Here we will reduce ourselves just to the 'What is AI?' question and will not deal with how to find such a program.

2. In [3] we defined AI as a program and here we will define AI as strategy. In [3] AI is a program and the world AI lives in is a program, too. Thus we end up with two programs playing against each other, which is somewhat perplexing. It is better to define AI as a strategy and have a program playing against a strategy. Certainly, the strategy will be computable, because it is finite. Our AI program will be any program which implements an AI strategy for the first 1000 games. What the AI program does after 1000 games will remain undetermined, however it is hoped that the program will continue to behave intelligently thereafter.

3. In [3] the world was presented by means of non-deterministic Turning machines. This is a futile complication. It would have made sense if there were no relation between the individual games and if the machine tape were erased (reset) after each game. And because the tape is not erased, each next game depends on what happened in the previous game (what has remained on the tape). For this reason we will use deterministic Turing machines which are simpler, while the fact that they depend on what has remained on the tape makes their behavior in the various games different (non-deterministic).

4. At each step we have Action and Observation. Actions and Observations in [3] consisted of a single symbol, while here the Action will consist of $n$ symbols and the Observation will be of $m$ symbols. We can certainly code multiple symbols into one, but this will be at odds with the idea that unnecessary coding should be avoided [6]. The world is complicated and hard to understand enough. Adding one unnecessary coding will make it even less understandable.

5. While in [3] it was assumed that all moves are correct, here we will add the concept of *incorrect move*. On one side it is important to assume that incorrect moves may exist. On the other side, this will spare us the indiscriminate shutdowns of the Turing machine, which we did in [3] in order to avoid cycling.

6. The Turing machine is a theoretical model which does not need efficiency. Here we will harness this model in real work and will therefore modify it in order to boost its efficiency. The complication of the model is the price to be paid for the so obtained higher efficiency.

7. In [3] we used the Turing machine in order to describe a logical world. If it is computable it should be logical. However, the world described by the Turing machine is not very logical. Everything is recorded on a single tape and the program is not structured at all. It jumps indiscriminately from one command to another ('spaghetti code', as software engineers have tagged this pattern). The way such a program operates is rather illogical and we will address this by adding subprograms and more tapes so the machine becomes a multi-tape one.

8. In [3] we defined IQ as an arithmetic mean which cannot be calculated precisely because of combinatorial explosion. What we say there is that it can be calculated approximately through a statistical sample. Here we will introduce the terms 'Global IQ' — which cannot be calculated precisely, and 'Local IQ' — which is readily computable through a specific statistical sample. The Local IQ will approach to the Global IQ when the size of the statistical sample approaches to infinity. The set of worlds we use for calculating the Global IQ is finite (enormous, but still finite). Nevertheless, the size of the statistical sample can tend to infinity because there may be repetitions in the sample (although repetitions are unlikely due to the hugeness of the set from which the sample is recruited).

## Related work

In [7], Turing proposes his definition of AI (the "Turing test"). The idea is that if a machine can successfully imitate a human, then this machine is AI. In the Turing test, as in our article, we have an exam and in order for a machine to be recognized as AI, it has to pass the exam. One difference is that there is an interrogator there, while we have a test with fixed tasks. That is, the Turing test is subjective, which is why it is an informal definition of AI. However, the main problem of Turing's definition is not in its subjectivity and informality, but in the fact that it does not define an intellect, but something more. It defines an educated intellect. For an intellect to pass the Turing test, it must be educated. We can even assume it has an Anglo-Saxon education because if it does not speak English it would not perform well on the test.

Intellect and educated intellect are two different things, just like a computer without software and a computer with software are two different things. If you ask a mathematician what a computer is, he will answer you: "the Turing machine". If you ask the same question to a child, the child will answer: "A computer is something that you can play games on, watch movies, etc." That is, mathematicians perceive the computer only as hardware, while the child perceives the computer as an indivisible system of software and hardware. When Turing gave a definition of a computer (the Turing machine), he described only the hardware, but when he defined intellect, he described an indivisible system of intellect and education.

Despite the Turing test defines educated intellect, Turing understands very well the difference between educated and uneducated intellect. In the end of [7] he asked the question: "Instead of trying to produce a program to simulate the adult mind, why not rather try to produce one which simulates the child's?"

The AI definition we give in this article answers to this question. It does not include education and the tasks in the test do not imply any preliminary knowledge. That is, with every task we will assume that we are starting anew, and that in the course of solving the task we are learning, i.e. we find dependencies and learn from our mistakes.

In [8], McCarthy says that the distinctive feature of AI is "the ability to achieve goals in the world". We will not argue with McCarthy, but we will only clarify what he has said. We will specify what a world is and what a random world is and what are the goals that AI needs to achieve. On this basis, we will create an IQ test where programs that achieve more goals have a higher IQ.

McCarthy also says that there is no "definition of intelligence that does not depend on how it relates to human intelligence." Indeed, in this article we calculate the value of IQ without being related to human intelligence, but to say if a program is AI, we have to say what the minimum for the IQ is. For this minimum, we chose the number 0.7. This number is arbitrarily chosen and we will change it if this level turns out to be much lower or much higher than the IQ of the human. That is, in our definition human intellect is also used, but this is only for comparison and we use it only to determine one special constant.

Yet another question McCarthy asks is: "Can we ask a yes or no question? Is this machine intelligent?" The answer is "no" and we fully agree because we do not known what the minimum for the IQ is.

However, we will not agree with McCarthy's answer to his next question. He asks: "Do computer programs have IQs?" And he answers with "no". In this article, as in [3], we've showed that IQ can be defined for programs. With his answer McCarthy probably wanted to say that IQ tests for people are not suitable for computer programs. The test we offer is not for people but for programs.

In some articles (for example, in [9]), the question is how to create a program that can solve IQ tests designed for humans. In our article, the issue we are dealing with is the opposite one. Here we create IQ tests designed for programs (for strategies). The tests we will offer will not be suitable for humans, although it is possible for a human to learn to solve them after some training. The trained human will note down what has happened and will analyze to find dependencies. The untrained human will not note down and will not be able to notice the dependencies unless they are presented in visual form or in other way which is convenient for perceiving by a human.

It is very difficult to think of a human as a strategy for the following reasons. First, humans are not deterministic, that is, they do not implement a deterministic strategy. (A human can be seen as a not deterministic strategy or as a set of strategies from which one is randomly chosen.) Secondly, it is unclear how much time we have given the human to make a move. Third, it is unclear how seriously the human will approach the task (if he approaches more seriously, he will get a better result). Fourthly, humans are always trained and never start from scratch. Even a newborn baby has gone through some training in the mother's womb. Therefore, when a human implements a strategy, the outcome largely depends on his education, training, experience.

In [10], Detterman threatens to "develop a unique battery of intelligence tests" that will be able to measure the IQ of computer programs. In this article, as in [3], we do not threaten, but we are creating such a test.

It is a bit confusing that Detterman uses "computer" when he means "the system comprising of a computer and a program". It is better to call this system a program because when we know the program, it's not very important which computer we'll run it on because the difference between two computers is only in their speed. Conversely, if we run two different programs on one computer, the difference in their behaviour will be enormous.

Detterman intends to test computer programs by using IQ tests for people. That is, he, just like Turing, does not make a distinction between intellect and educated intellect. Therefore, Detterman's test will not provide time to learn, but will assume that the computer program that is being tested has been pre-trained.

Detterman relies on the fact that computers are better than people in finding factual information. This is an ability of computers that is not directly related to the intellect. Similarly, computers are very powerful in arithmetic operations, but that does not make them smart.

There is one thing in which we agree with Detterman. He notes that with hoc algorithms many tasks can be solved, but the real intellect must be able to find the solution on its own.

In [11], the authors set the ambitious task of developing an IQ test that is appropriate for both AI and humans, and for programs like Siri and AlphaGo that are not AI. We can not compare AI with programs that are not AI because the first is a program that can be trained for a random task and the second is a program written for a specific task. For example, how do I compare a chess program with AI? The only thing a program designed to play chess can do is to play chess while AI can do everything although not immediately but after being trained. That is, the only way to compare these two programs is to let them play chess, and the chess program will have an advantage because AI will waste time learning, while its opponent would not need to learn how to play chess because it knows how to do it. If we compare the chess program with AI which is trained to play chess then the first program would still have some advantage over AI, just like the specialized hardware (i.e., hardware made specially designed for a particular task) takes precedence over a computer program running on a computer. That is, the idea of comparing AI with programs that are not AI is not good.

The authors of [11] refer to concepts such as: abilities to acquire, master, create, and feedback knowledge. It is like explaining a term whose meaning we do not know by means of other terms whose meaning we do not know.

[12] discusses very comprehensively and competently how AI's IQ is measured. Also in [12], a very good overview of the different works devoted to this topic has been made. One minor disadvantage of [12] is that there is some controversy there. On the one hand, the article says "IQ tests are not for machines yet". On the other hand, in earlier articles by Orallo [13, 14] a formal definition of AI based on the IQ test is given. It seems like Orallo is taking a step back and relinquishes the previous results. Another contradiction in [12] is that, on the one hand, it says "human IQ tests are not for machines". This is something we fully agree with, as we agree with the arguments that accompany this statement. On the other hand, in the same article, the authors say they agree with Detterman that "there is a better alternative: test computers on human intelligence tests".

In [12], as well as in [11], a universal IQ is sought, which is applicable to all programs and even to humans. Here we differ from the authors of [11, 12]. We have already explained why it is not a good idea to compare AI with programs that are not AI. We also explained why it is not a good idea to use the same tests for AI and for humans.

Article [13] is of significant importance because this is the first article that talks about a formal definition of AI, and this is also the first article introducing the IQ test for AI. Indeed in [13] this test is called C-test, but it is explicitly said that it is an IQ test for AI. We have to apologize for omitting to quote [13] in [3]. This is an omission that we are now correcting.

Despite the seriousness and comprehensiveness of [13], there are some inaccuracies that we have to mark. The fact that we pay attention to some omissions and inaccuracies in [13] does not in any way mean that we underestimate the significance of this article. There are no perfect articles, and inaccuracies can be found in any article. Usually the first article that appears in a new area is slightly confused and unclear. Normally, in later articles, things get clearer and more precise.

The most serious inaccuracy of [13] is that it does not define AI but something else. We will call that another thing an "observer". We could say that an observer is a program that has only input and no output, but that would be too restrictive. That is why we will say that an observer is a program whose output does not affect the state of the world but the output can influence the reward that the program will receive.

An example of an observer is when we play on the stock market with a virtual wallet. We do not influence stock prices because we do not play with real money. (Even if we played with real money, it can still be assumed that our actions do not affect the stock exchange, because when playing with small amounts our influence is negligible.) When playing with a virtual wallet, at every step we change our investment portfolio and after each step the value of the portfolio changes depending on the change in stock prices. The reward we receive will be our virtual profit or loss.

The fact that the program defined in [13] can not influence the world is substantial problem, because as people say: "You have to touch to learn". There are other proverbs that say that we can not learn by watching only. By depriving AI of the opportunity to experiment, we are seriously restricting it.

In [5] it is noted that there are two problems that AI needs to solve. The first is to understand the world (to build a model of the world), and the second problem is to plan its actions based on the chosen model. That is, the "observer" solves only the first problem without solving the latter.

In [13] the definition is limited to "observers", but to give the term "observer" a definition is a task sufficiently meaningful by itself because it is half of the things AI needs to do. Unfortunately, this task is not fully solved because the observer defined in [13] is a little more special. It either understands the world wholly or not at all. That is, it is observer who works on the principle "all or nothing".

[13] suggests random strings, which can be continued in only one possible way. That is, it is supposed that there is a single simple dependence and this dependence will either be found or will not be found. This approach is too restrictive because it implies only observers who understand the world completely. This is only possible in very simple worlds. Any more complex world can not be fully understood and that's why it has to be understood only partially.

The authors of [13] are making serious efforts to make the dependencies exceptions-free. The definition of exception-free given in [13] is very interesting. However, we would not go that way because if dependencies included exceptions, that would be a way to allow a partial understanding of the world. Seeking a single exception-free dependence to describe everything is the reason why the observer defined in [13] relies on the "all or nothing" principle.

We have a few other minor recommendations for [13].

We give some time to AI to find dependence. The question is whether we will give this time at once or we will distribute it in many steps. In general, AI is searching for dependencies throughout its entire live. That is, in many steps. In [13], dependency is sought in just one step. In our article, we assume that in life the steps are approx. one million (a thousand games with a thousand steps each one). This means that we must give the program, which is defined in [13], a million times longer time. Our recommendation is for dependences to be sought after each step, not only after the last one.

In [13] we do not like also the fact that the program that generates the test is not working due to combinatorial explosion. This is the program called "The Generator". That is, [13] does not present a test, but only shows that such a test exists theoretically.

Yet another problem is that two restrictions have been imposed. When the "observer" makes a prediction, it must bet all on single result only and always bet the same amount. It would be better to have the freedom to bet on more than one result, and to decide what amount to bet. When we are more confident we bet more, and when we hesitate, we bet less or we pass. These two limitations do not change things fundamentally, because the smart will prove to be smart even with these limitations, but that blurs the picture. If there were no such restrictions, the difference between the IQ of the clever and the stupid would be greater.

We also do not like the fact that the more complex tasks in the [13] have a greater weight than the simpler ones, when it should be the opposite. (There is a coefficient $e$, which is assumed to be non-negative and would be better to be negative.) It is true that when we make a test we give more points to the more difficult tasks, but that's because we suppose that students will lose more time on the more difficult task. This is not the case here. Here, for every task we give the same time. Therefore, if a simple task can not be solved, it is a serious problem and this should be reflected in the score. Moreover, sometimes we will guess the solution of a difficult problem randomly, so those must be given less weight, otherwise they will result in undeserved rise in the IQ.

We have the concepts of global and local IQ (these two concepts are defined in our article). Global IQ is something accurate that can not be accurately calculated (because of a combinatorial explosion). Local IQ is not something accurate because it depends on the choice of the specific tasks in the test, but for specific tasks it can be easily and accurately calculated. What is defined in [13] is local IQ, not global IQ. However, let's not forget that the local IQ approaches the global IQ, but [13] says nothing about the "nerd" program, which is the main problem of local IQ.

How are we to correct the definition given in [13] in order to get a definition of the observer that does not follow the principle "all or nothing"?

Instead of taking a special k-comprehensible sequence, we will take a totally random program and the sequence that this random program generates. Instead of predicting the next symbol a single time, we will predict it at every step. We will not bet everything on a single prediction, but we will allow the bet to be divided between several predictions. We will also allow the bet to be different. (For example, we will assume that the sum of bets for the last five steps is limited to a given constant, but that we have the freedom to choose when to bet.) The reward to a sequence will be the sum of the rewards of all steps. Local IQ will be the arithmetic mean of the rewards to the sequences we have included in the test.

In this way, we will not want the "observer" to understand the world completely because it can grasp some dependencies and thus, through a partial understanding of the world, get a relatively high IQ.

A problem with [13] is that the program it defines is actually the program [1]. This is a simple program that predicts the next symbol on the basis of the simplest dependency that can generate the beginning of the sequence.

I even dare go a bit further and say: No program satisfying the definition given in [13] is better than [1]. That none is better as a result is clear, but I say that even as a matter of efficiency none is better because the definition in [13] does not give us any opportunities to play tricks and discover dependencies partially (step by step), leaving us the only option to foolishly go through all possible dependencies.

Interestingly, in his more recent articles Orallo (as in [14]) has obviously understood the main omission made in [13], and what he defines there is no longer an "observer" but a program that can influence the world. Unfortunately this program is not AI again. [14] defines a program that plays a game which consists in going round a labyrinth (graph) by chasing something good and running away from something bad. To play this game one definitely needs intelligence, but the program playing this game is not AI just like the program playing chess is not AI.

Again in [14] the authors speak of reinforcement learning. That is, it is clear that they know very well what the general form of AI is. Why, then, do they not use this general form, but restrict themselves to the worlds of one specific game? I think the reason is that they are trying to avoid the worlds where fatal errors are possible. The problem with fatal errors is mentioned even in [2], but fatal errors are not really a problem. We humans also live in a world with fatal errors, but that does not prevent us showing who does better and who does worse. Well yes, it's a problem for people because we live only one life, but the IQ test consists of many tasks, each of which is a separate life. Even if fatal errors occur in several lives, this will not significantly affect the average score. Even with humans, life is not just one. From the point of view of the individual, life is one, but from the point of view of evolution, lives are many. Some of our heirs will die because of fatal mistakes, but others will survive. Thus the average score of our successors will not be significantly affected by the fact that some of them have made fatal errors.

It is true that in this article, as in [3], we also do not use a random world, but we only take computable worlds, but this constraint is not essential, because every time-limited world is computable, and we can easily assume that all worlds are limited in time. That is, by limiting ourselves to the computable worlds, we are not limited at all, but only give more weight to the worlds that are simpler (according to Kolmogorov complexity).

In [14] there is one more thing we do not agree with. This is something called the "discount-rate factor". Of course, this is not something conceived by the authors of [14], but is something widespread among people working in the field of reinforcement learning. The idea of the "discount-rate factor" is that the past is more important than the future. Life is potentially endless, and to evaluate an endless life, we have to devalue the future. However, the past being more important than the future is not a good idea. It would be better to do the opposite, because in the past we have not yet been trained, and in the future we have already been trained. With humans, we do not count how many times a person wet his bed while he was a baby. Instead, we see what achievements the person has attained in his adult age. Therefore, the approach we have adopted in [3] and this article is that there is no "discount-rate factor" but life is limited. In other words, the approach here and in [3] is that the discount-rate factor to be equal to one until a given moment in time (the end of life) and to be zero from that moment on.

# Formulation of the problem

Let us have a Device which lives in a certain World. At each step the Device produces *n* symbols (this is the Action) and then receives *m* symbols from the outer world (in our terminology the first one will be 'Reward' and the remaining *m-1* symbols will be 'Observation'). The Reward can have five possible values: *{nothing, victory, loss, draw, incorrect_move}*.

We will use the words 'move' and 'action' interchangeably. If we perceive life as a game it is more appropriate to say *move* rather than *action*. Likewise, we will use the words 'history' and 'life' as synonyms.

Let one step of the Device be a triple consisting of <Action, Reward, Observation>. The 'life' of the Device will be a sequence of steps resulting from the interaction of the Device with some World.

'Real life' will be life without incorrect moves. Therefore, all <Action, Reward, Observation>. triples where the Reward value is '*incorrect_move*' must be removed from the life and what remains is the real life.

A 'moment' will be a sequence of steps where the last step before the sequence and the last step of the sequence are correct and all steps in between are incorrect. That is, in *life* there may be more steps than moments, but in *real life* the number of steps is equal to the number of moments.

We will assume that the behavior of the Device and of the World is deterministic. In other words, we will assume that if we know which is the Device and which is the World then we know which is the history.

The behavior of the Device can be presented as a strategy, i.e. as a function which defines the next move of the Device for each start of life. Likewise, the World's behavior can be presented as a strategy which for each start of life and for each Action of the Device delivers the Reward and the Observation which the Device will get at the next step. It should be noted that the World's strategy does not depend on incorrect moves. Hence, we can imagine the strategy of the World as a function of real life. Conversely, the strategy of the Device will depend on incorrect moves (these moves will provide additional information for the Device to use).

The Device and the World can be thought as two strategies playing against one another, but that would not be accurate, because the Device has an objective and the World has not. Therefore, the World does not play against the Device. We assume that the World is simply there and does not care whether the Device feels good or bad.

Presenting both the World and the Device as strategies is not a very good idea, because a strategy remembers everything (i.e. depends on entire life until the moment). It makes sense to assume that the Device may not necessarily remember everything. A similar statement can be made in respect of the World. There may be a world the entire past of which (previous life until the current moment) can be reconstructed out of its internal state. It may be, however, that the world does not remember everything, which leaves us that two different histories can lead to the same internal state of the world. This is the reason why, we will present the Device and the World as functions.

Let us have two sets, *Q* and *S*. These will be the sets of the internal states of the Device and of the World. These sets will be finite or countable, at the most. Let $q_0$ and $s_0$ be the initial states of the Device and of the World. We will assume that these states are fixed, because life will depend on the initial states we start from, but we want life to depend only on the Device and on the World.

The Device and the World will be the following functions:

*Device: Q×Rewards× Observations×$2^{Actions}$ → Actions×Q*
*World: S×Actions → Rewards×Observations×S*

For each Internal State of the Device, Reward, Observation and Set of Moves which are confirmed as incorrect at the moment, the *Device* function will return an Action and a new State of the Device. We will assume that the *Device* never returns an Action which is confirmed as incorrect at the moment.

The Internal State of the Device will reflect everything the Device has remembered. What can it have remembered? It can remember everything that has happened until the moment plus its last Action. That is why our expression is *Actions×Q*, rather than *Q×Actions*. We wish to stress that the Internal State of the Device can remember the last Action.

For each Internal State of the World and Action, the *World* function will return a Reward, Observation and a new Internal State of the World. It makes sense to assume that there are moments (Internal States of the World) in which a specific action is impossible or incorrect. It is natural to assume, therefore, that the *World* function is a partial one. We will supplement the definition of the function so as to accommodate those moments as well, and in this way will extend the function to a total one. In these moments *Reward* will equal *incorrect_move*, the *Observation* value will be irrelevant and the new internal state will be the same as the previous one although this is irrelevant, too.

The Internal State of the World will reflect what the World has remembered. What can it have remembered? It can remember everything that has happened until the moment plus the last Reward and Observation. That is why our expression is *Rewards×Observations×S*, rather than *S×Rewards×Observations*. We wish to stress that the new Internal State can remember the last Reward and Observation.

In [2] and [3] we defined the new <Reward, Observation> as a function of the new internal state. So we assumed that they *must* be remembered, but now we dispense of this requirement. Let us take a world in which we play chess. The chess game ends and the new internal state of the world is a chessboard with the initial lineup. In this case we do not have to remember who won the last game. Such requirement would be one unnecessary care.

This is how the life of the Device looks like:
<$a_1$, $r_1$, $o_1$>, <$a_2$, $r_2$, $o_2$>, …, <$a_{t-1}$, $r_{t-1}$, $o_{t-1}$>

Let us see how the *Device* and the *World* function define life.

<$a_{i+1}$, $q_{i+1}$>= *Device*($q_{i-j}$, $r_{i-j}$, $o_{i-j}$, *incorrect_actions$_i$*)
<$r_{i+1}$, $o_{i+1}$, $s_{i+1}$>=*World*($s_{i-j}$, $a_{i+1}$)

In this case, *i-j* is the last correct step before *i+1*. The *incorrect_actions$_i$* set contains the confirmed incorrect actions at this moment. The set has *j* elements. Therefore, *incorrect_actions$_i$* *={a$_{i-j+1}$, …, a$_i$}*. The *incorrect_actions$_0$* set will be an empty set, because there will not be any confirmed incorrect actions in the first moment before the first step is made.

In order to define life, we must fix the first Reward and the first Observation ($r_0$, $o_0$). We do not wish life to depend on the first Reward and the first Observation, so decide that the value of all symbols of these two vectors is *nothing*. This is a sensible decision, because it is quite natural that in the first moment we do not get any reward and do not see anything significant (i.e. what we see in the first moment is the zero step). We will not define the zero action $a_0$, because we do not use it.

Now that we have fixed ($q_0$, $s_0$, $r_0$, $o_0$, *incorrect_actions$_0$*) we can build life up to step *t* and this life will depend only on the functions *Device* and *World*.
$<a_1, r_1, o_1>, <a_2, r_2, o_2>, …, <a_{t-1}, r_{t-1}, o_{t-1}>$

In addition to life, we will also construct the series of internal states of the Device and of the World, as well as the *incorrect_actions$_i$* series of confirmed incorrect moves at the relevant moment. Let us note that at each given moment there may be many incorrect actions, but the confirmed ones are only those which we have tried and have thus verified that they are indeed incorrect at that moment.

We will assume that the *Device* function always returns a move which is not confirmed as incorrect. The opposite will lead to cycling. What shall we do if all moves turn out to be confirmed incorrect moves? (I.e. if *incorrect_actions$_i$* coincides with the entire *Actions* set.) In this case we will assume that the *Device* function is not defined. This is the case when we end up in a blind alley and there are no possible next moves.

**Note:** The definitions of *Device* as a function and as a strategy are almost identical save that if we consider *Device* as a strategy, the order in which we tried the incorrect moves may matter at some moments, while in the function definition it would not matter. This difference can be resolved in two ways. The first one: when defining a function, instead of the set of confirmed incorrect moves we take the list of these moves. The second one is to limit our exercise only to those strategies which are insensitive to the order in which we tried the incorrect moves. In this study it would not matter what happens if the strategy tries the incorrect moves in another order, because we assume the strategy to be deterministic and the order in which it tries the incorrect moves is fixed.

**Note:** We assume here that when we try to play an incorrect move nothing happens and all we get is information that the move has been incorrect. We can let the Device check whether a move is correct or not even without necessarily playing that move (as we have done in previous studies). In this case we need to modify the formulation of the problem and add an additional reward *correct_move*. The *Device* function will need one more argument containing the confirmed correct moves. When we try a confirmed correct move we will assume that we shall play this move. Where a move is not in the set of confirmed correct moves, then we assume that we only try it. The *World* function will have another Boolean parameter to indicate whether the move is actually played or just tried. In the latter case, the function will return either *correct_move* or *incorrect_move* rewards. Then the move will not be played but will only be added to the next set of confirmed correct or confirmed incorrect moves.

A game will be a stretch of life located between two consecutive final rewards. In our terminology, 'final rewards' stands for the values *{victory, loss, draw}*. We will assume that the length of each game is limited to 1000 moves (i.e. 1000 moments, while the number of steps may be greater because of incorrect moves).

We will define which strategy is an AI strategy and our definition will depend on a number of parameters. We will fix the values of the majority of these parameters to one thousand, because 1000 is a nice round number. Another good round number is one million. Replacing one thousand with one million would produce another AI definition, which will not be much different from the one we are dealing with.

## Parameters

| Number of Action symbols | n |
|---|---|
| Number of Observation symbols | m |
| Number of symbols possible for each Action symbol | $k_1, \ldots, k_n,$ $k_i \geq 2.$ |
| Number of symbols possible for each Reward and Observation symbol | $k_{n+1}, \ldots, k_{n+m},$ $k_i \geq 2, k_{n+1}=5.$ |
| Tape symbols count | $MaxSymbols = 10 + \max_{i \in [1,\ n+m]} k_i$ |
| Global tapes count | 7 (from 3 to 9) |
| Internal states count | 1000 |
| Test worlds count | 1000 |
| Maximum number of games per life | 1000 |
| Maximum number of moves per game | 1000 |
| Maximum number of Turing machine steps per one life step | 1000 |
| Probability by means of which the machine is generated | $\frac{1}{10} = 10\%$ |
| Minimum IQ required for a strategy to be recognized as AI | $0.7 = 70\%$ |

The first four rows of the table above provide the parameters which describe the AI input and output. They tell us what the format of the sought AI is. Thus, we cannot vary these parameters randomly. The next eight parameters influence the choice of worlds selected for the test and, therefore, influence the IQ we get, and thus the AI definition as such. The last parameter also influences the definition. This means that if we varied the last nine parameters we would also vary the AI definition.

The table does not include other parameters which also influence the AI definition. For example, it does not say which pseudorandom numbers generator we will use in order to select the test worlds. Thankfully this parameter's bearing on the definition is insignificant.

The possible symbols for the *i*-th Action symbol will range from 0 to $k_i$-1. Likewise, the possible symbols for the *i*-th Observation symbol will range from 0 to $k_{n+1+i}$-1. Let us call the 0 symbol 'nothing'. The first Observation symbol will be the Reward. When it comes to the Reward, we will name the symbols 1, 2, 3 as 'victory', 'loss', 'draw', and these will be the final rewards. The reward 4 (incorrect move) will not be returned by the Turing machine as a result of the invocation of the command $q_1$. This reward will be returned only when Turing goes cycling (makes more than 1000 steps without reaching a final state) or crashes (e.g. invokes a **return** command when the stack is empty).

The tape symbols will be as many as needed for coding the Action and the Observation. I.e. the maximum $k_i$ for $i$ from 1 to $n+m$. To that we will add another 10 utility symbols the first of which will be empty symbol λ.

## How will the test work?

We will select 1000 worlds for the test. In each of these 1000 worlds, the candidate will live one life consisting of not more than 1000 games. Finally we will calculate the number of victories, losses and draws. This will give us an IQ, which is an arithmetic mean where victory is 1, loss is 0 and draw is 1/2.

We will pick the worlds randomly, but we want the selected worlds to be fixed, so we will use pseudorandom selection by setting the pseudorandom generator to 1 before starting the selection process. Thus we will always use the same worlds in the test.

In many of the generated worlds we will win every time or we will loss every time (doesn't matter what we do). It is meaningless to include such worlds in the test so we will discard all of them from test. Thus we will be left with 1000 meaningful worlds.

## What is a World?

In newspapers one can come across problems such as 'Which is the next number in the series?' If all possible series are equally probable, the next number may be any one. When looking for the next number in a series we assume that simpler series are more probable than more complex series. We therefore use the principle known as 'Occam's razor'.

The situation with worlds is similar. If we perceive the world as a strategy and all strategies are equally probable, then there is no way we can tell the future and a basis for preferring one move over another is lacking. We will therefore apply Occam's razor to the worlds and will assume that more simple worlds are more probable. What makes one world simpler than another world? We will use the Kolmogorov complexity. That said, if the world is a strategy, the more simple strategy is the one which is generated by a Turing machine with less states.

We have limited the life so all strategies are finite. Therefore, all strategies are computable. So we will assume that a world is some Turing machine which calculates some strategy.

# Which will be the set of all worlds?

The set will be limited to Turing machines with 1000 states. This set includes machines with less than 1000 states, because each machine can be filled up with unreachable states. Turing machines which use more than 1000 states (and cannot be simplified) will be excluded from the set. We will consider these worlds as too complex and will accordingly exclude them from the definition.

The result is a huge set of strategies (worlds). The more simple strategies here will bear more weight (will be more probable), because there are generated by more Turing machines (from the set).

Furthermore, we will assign some weight to each Turing machine. Hence we will prefer some machines over others. For example, if a machine tends to use more states with lower numbers, we will prefer it to the one which uses more states with higher numbers.

There are two reasons why we assign different weights to the different Turing machines. The first one is that we give more weight to more simple machines (e.g. the ones with less reachable states are simpler). The second reason is that we wish to randomly generate a machine which works, and this is very difficult. The machines which have a greater chance of being working ones will have more weight, therefore we increase the probability of selecting these machines and accordingly our chances of hitting a working machine.

The weight of a machine is equal to the probability of this machine to be selected by us. We will not calculate this probability. That would be a rather difficult calculation. We simply select a machine randomly and by this selection we induce probability as a parameter of the formula. That is, the probability will not be calculated when the Local IQ is calculated. When calculate the Global IQ, we should consider all Turing machines in order to calculate the success rate of the Device with any of these machines and the probability that this very machine is selected. We have to multiply these numbers and sum up the values for all machines. This is an impossible calculation because of combinatorial explosion. The fact that we complicate the calculation by adding the calculation of probabilities is not a problem. We do not change anything in this way. While the Global IQ remains theoretically computable, in practice it is still uncomputable, even more uncomputable than before.

# How to calculate the IQ of a particular computer program?

We may say that the IQ is the arithmetic mean of success rates of all worlds. (Note that the worlds are not equally probable and we need to multiply the success rate by the probability (weight) of each world.)

We will call the so obtained IQ a Global IQ. The Global IQ definition is very nice save that Global IQ cannot be calculated. To be precise, it can be calculated in theory, but not in practice because of the huge number of worlds which we have assumed to be possible.

Nevertheless, we can still calculate the Global IQ by approximation using statistical methods. We will select randomly 1000 worlds and will calculate the arithmetic mean for these worlds. The result obtained would be close to the Global IQ.

The problem with this approach is that different selections of test worlds will yield different Global IQ approximations. What we need is a program which awards to the candidate the IQ this candidate deserves, withal it must be a specific value rather than an approximation of something else. Hence we will fix the randomly selected 1000 worlds and will say that the Local IQ is the average success rate across these 1000 worlds. (In this case different worlds will not have different weights, because the weight is already accounted for in the selection of test worlds. The more weighty ones are more likely to be selected.)

The idea of fixing the randomly selected worlds is tantamount to giving the same problems to all candidate students.

The Local IQ is an easily computable function and describes well our understanding of what IQ is. There is just one problem. There is one program which we will call a 'crammer'. This program is designed specifically for the 1000 test worlds and its Local IQ is very high, but its Global IQ is low. How should we resolve this issue? We will use the Local IQ in order to find the AI. When we hit a program with a very high Local IQ and suspect that this program is a crammer, we will give it additional problems, so that we calculate a second local IQ. This means that we will take the next 1000 random worlds and derive another arithmetic mean from these worlds. We can go on with a third and fourth Local IQ.

## How does Turing machine make a move?

We present the world as a Turing machine. Thus, our machine must take Actions as inputs and deliver Observations as outputs.

The first $m+1$ states of the machine will be special. The $q_{m+1}$ state will be the initial and the final state of the machine. In states $q_1$ to $q_m$ we will use in order to output the Observation symbols.

When the machine makes its first move, all tapes will be empty (i.e. they will be covered with the λ symbol). The first running tape will be No 3 (Nos. 0, 1 and 2 are reserved for utility purposes).

Turing machine will make a move by starting from the initial state $q_{m+1}$ and finishing in the same state (which is also the final state). In the beginning of the move, a word of $n$ symbols (the Action word) will be recorded on the current tape under the head of the machine (anything previously written in the first $n$ symbols on the tape before the recording of the word will be deleted).

At every step it will be observed if the states $q_1$ to $q_m$ are called up. When these states are called up, the Observation symbols will be outputted. If the $q_i$ state is called several times within one move, only the first call will be considered. If it is not called up at all, the $i$-th Observation symbol will be *nothing*. If it is called up at least once, $i$-th Observation symbol will be the value of the 'head memory' at the moment following the first call on the $q_i$ state. If the $i$-th Observation letter is too big (equals or exceeds the $k_{n+i}$), the machine will crash in a cycling-like fashion.

# Incorrect moves

When the Turing machine fails to make a move because it goes cycling or crashes for some other reason, we will deem that the input Action entered at the start of the move is incorrect or impossible. In this case we go back and try to enter another Action. More precisely, it is not us who go back, instead we will turn back the world (the Turing machine). We give the Device an *incorrect _move* reward and take the next move the Device chooses to give us. We continue with this move as if the Device has played it instead of the incorrect move. (If the Device plays the same incorrect move, we disqualify it because it cannot try the same move twice in a single moment.)

Taking the Turing machine back is a perfectly natural operation. All we need is to memorize the total state of the machine before the start of the move. If we reconstruct this total state, we can enter another Action and continue was if it were the first action we ever tried.

Another approach was applied in [3], where it was assumed that all moves are correct and the cycling problem was resolved by aborting the execution of the program, awarding a utility draw and restarting the program from its initial state. With this type of restart the tape remains in the state it was at the time of abortion. This is very poor practice. You will know that to turn down your PC you should enter the 'Shut Down' command. The other option is to pull the plug off the socket, but this will leave the hard disk in the state it was when the plug was pulled off. This treatment would make your PC behave in a bizarre and illogical manner. The same can be said when a Turing machine is shut down randomly and then restarted from its initial state. We want the world to be as logical as possible and take care to avoid these abortions.

Now that we have opened the gate for incorrect moves, we should know what to do when all moves are incorrect. Let us assume that we have about 100 possible Actions. We try them all and all of them prove to be incorrect. So we will say that we have ended in a blind alley with no way out.

If life is the sequence of 1000 games, then life ends with a natural death after 1000 games or with a sudden death (blind alley). How to reward a life that ends prematurely? We can reward only the games played until death. Then our AI strategy will prefer to commit a suicide (enter a blind alley) as soon as it realizes that things in the current life have gone wrong and only losses lie ahead.

We do not want suicidal AI strategies and will therefore opt for another solution. When a strategy is caught in a blind alley, we shall say that all remaining games up to 1000 are losses. This will provide assurance that the strategy will not go in blind alley on its own will, but will keep fighting to the end.

Will the strategy realize that it is entering a blind alley? One cannot learn this by trial and error because you fall in a blind alley just once. But, if the strategy is very smart, it may be able to predict that some blind alley is forthcoming. For example, if the number of possible moves is decreasing, this is not a good sign, because in the end of the day one may run out of possible moves and enter in a blind alley. People can be another example. Let us have someone who has never died. He or she has no experience of their own but can still predict a few potentially fatal situations.

## We penalize beating around the bush

We said that a game cannot be longer than 1000 moves. What shall we do if a game continues for more than 1000 moves? We shall then award a utility draw. Note than by doing so we do not interfere in the workings of the Turing machine as it continues to play the same game. The intervention concerns only the strategy, because it will get a *draw* reward when the world (Turing machine) has awarded *nothing*. The fact that we do not abort the machine guarantees that the machine will maintain its logical behavior.

If another 1000 moves are played without a final reward, we award a utility loss. That is, we penalize strategies which kick the can down the road. We want an AI strategy which aims to close the game quickly and start the next one.

By penalizing early death and procrastination we vary the IQ of the random strategy. If we play heads or tails, the expected IQ is 1/2. To earn a higher IQ one must purposefully aim to win. For a lower IQ, one must purposefully aim to lose. By declaring all post-death games lost, we reduce the IQs of all strategies. Similarly, the inclusion of utility losses will lead to a likewise reduction. This solution means that the IQ of the random strategy will be less than 1/2.

We will be able to calculate the exact IQ value of a random strategy once we write the program which calculates the Local IQ. Naturally, a random strategy is non-deterministic meaning that we should test it a few times and take the mean value. Hence, the IQ of the random strategy will be an approximate value. Only deterministic strategies have exact Local and Global IQ values.

## More logical and more efficient Turing machine

As announced already, we will redefine the Turing machine in order to make it more logical and more efficient. To this end, we will make Turing a multi-tape machine and will enable it call subprograms.

Why would such a machine deliver a more logical world?

First, it is the multiple tapes of the machine. It is legitimate to present the state of the world as a Cartesian product of many parameters which are in weak correlation with one another. That is why a multi-tape Turing machine produces a more logical model of the world than the single-tape machine.

Second, the machine will have subprograms which can be called up from many points. This is more logical than calling a different subprogram each time. When no stack is in place, we have to remember where we should go back after the subprogram is executed. To this end, each subprogram should be called up from one point only (otherwise it will not know where to go back).

When we invoke a subprogram we give it a clean tape so it can write its intermediate results. In the opposite case the subprogram will have to use for this one of the common tapes and this will largely frustrate its logic due to the weird interactions that will occur among different calls of the subprogram.

Why would this machine make less steps and have less internal states?

Higher efficiency means less steps and, more importantly, less internal states. It is important that the steps are not too many, because a machine making more than 1000 steps per move will be presumed to be in a cycling situation and will be stopped. It is also important that internal states are not too many, because we have limited our set to machines having less than 1000 states. Hence our machine should preferably use less states.

Having multiple tapes in the machine means there will be less steps, because a single tape will force the head make a lot of movements in order to write intermediate results. It would be easier to write these results on another tape.

More significantly, we will reduce the number of internal states of the machine. A classical Turing machine uses a huge number of internal states (it memorizes everything that needs to be memorized in its internal state). For example, when a subprogram has been called up we must remember where the subprogram was called from (where it must return). If we want to move a symbol from one place to another we must remember which symbol we have pulled up.

Thus, we complicate the Turing machine so that, in addition to its internal state, it should also remember which is the running tape, the stack index (for subprograms) and a 'head memory' symbol.

## A stacked Turing machine

How would the program of this machine look like? It will be a table of 1000 x MaxSymbols, where 1000 are the possible commands and MaxSymbols are the possible symbols. In each field of the table there will be five commands.

The first command is 'Write on tape'
MaxSymbols+2 possible values:
(unchanged, previous value of the head memory, concrete symbol)

The second command is 'Change the head memory'
MaxSymbols+2 possible values:
(unchanged, previous value of the symbol on tape, concrete symbol)

The third command is 'Move head'
Three possible values:
(left, right, stay there)

The fourth command is 'Subprogram'
There are two fields:
'State', the value of which is in the interval [0, 1000] (where 0 is the command *NULL*).
'New current tape', the value of which is in the interval [0, 9] (where 0 is the previous current tape, 1 is the current tape of the parent program, 2 is the temporary tape created especially for this invocation of the subprogram and 3—9 are the global tapes).

The fifth command is 'Next state'
The value is in the interval [0, 1000] (where 0 is the command *return*).

When a subprogram is called up, what do we write in the stack? Three things: where shall we go back after *return* (the fifth command), which is the previous current tape (so that we recover it after *return*) and which temporary tape is created especially for this invocation of the subprogram. The new temporary tape should also be written somewhere. Let it not be in the stack, but somewhere else. When the *return* command is executed, the corresponding temporary tape will be destroyed.

## Populating the table

In order to create a random Turing machine, we must populate 1000 x MaxSymbols with random commands. Before we do that let us say how a random command is selected. We must generate 6 numbers (the fourth command has two fields). These numbers may be equally probable, but we prefer smaller numbers to be more probable than larger numbers. Why would we prefer so? Because if all states are equally probable, the program will be stretched over multiple different states, while we wish certain states to be used more frequently than others. Likewise, we want some tapes to be used more often than others. This also applies to utility symbols (but does not apply to non-utility ones).

How shall we select a number from 0 to k with a decreasing probability? Let us toss a coin and if it is heads we select with a probability of 1/2 the number 0, if it is tails we toss again and if we get heads we select with a probability of 1/2 the number 1 and so forth. If do not get any heads all the way to k, we restart from 0.

A probability of 1/2 produces a very steep decrease of the probability of the next number. That is why we will use a probability of 1/10 to select all the 6 numbers. What we actually use is the geometrical distribution.

**Note:** Only the number of the subprogram will be assigned differently. In this case 0 will be taken with a probability of 9/10 instead of 1/10. (Thereafter we continue with 1/10.) This is so because we do not want subprograms to be invoked too often and the stack to be populated redundantly. In this way the probability of the *return* command will be equal to the probability of invoking a subprogram.

Now that we have explained the generation of commands (one box in the table), let us explain the generation of one column consisting of MaxSymbols boxes. We will imagine the command of this machine as a switch with MaxSymbols cases. When we write program and use a switch, we describe only some of the cases rather than all cases. The remaining cases are described as *default* (i.e. all other cases are the same). A program will be more logical if the same command applies to the majority of the cases. Therefore, we will select randomly how many different commands will exist in the column. We will use the decreasing probability procedure as explained above. Once we know how many different positions are there, we select randomly which these positions will be (again, smaller numbers will be more probable). Finally, we populate the positions that need to be different with different commands, and will put the same command in all other positions.

So we have an algorithm for populating one column and can populate 1000 columns. In the end of the process we have the first random Turing machine. This however is a cumbersome process, so we derive the second machine from the first one just by changing the first *m+1* states (which are special ones) and another 10 random states. This change is sufficient because the vast majority of the states are not used and by changing the special states we start using other states. Thus, the new Turing machine will be substantially different in terms of the states it uses even though the two machines are almost identical in terms of their unreachable states.

## Discarding the slag

So we have a fast and simple procedure to generate 1000 test worlds. The problem however is that most of these worlds are useless. Across 1000 worlds there may be as little as 2 or 3 interesting ones. Therefore, we will discard all useless worlds and do our test with 1000 interesting worlds.

Example for a useless world is if the very first move leads to a blind alley.

We will verify that a world is interesting by letting the random strategy live one life in that world. In this life, we do not want to see the strategy caught in a blind alley. We want at least one victory and at least one loss. We do not want more than 10 utility draws and utility losses. If all these conditions are met in this random life, we will deem that the world is interesting.

How to construct a test with 1000 interesting worlds? First, we initialize the pseudorandom numbers generator with the number 0. Then we generate randomly World Zero. Then we initialize the generator with 1 and apply a minor change to World Zero so as to generate World 1. If World 1 is interesting, we set the generator to 2 and generate World 2. If World 1 is useless we initialize the generator with 2, 3, 4, 5, etc. until we obtain from World Zero an interesting World 1. We keep going until we generate 1000 interesting worlds. We will not save them all, but will only save an array of 1000 numbers. These are values which we should use to initialize the generator in order to obtain a new interesting world from the previous interesting world.

**Note:** Please observe that we select different Turing machines with different probabilities. These different probabilities are the different weights of the different machines. We need this clarification, because we aim to provide an accurate definition of Global IQ. The definition is:

$$\text{Global IQ}(Strategy) = \sum_{TM}^{TM \,\in\, Interesting} P(TM \mid Interesting) \cdot \text{Success}(Strategy, TM)$$

In this equation, P(TM | Interesting) is the conditional probability that the machine named 'TM' is selected if the world is interesting. Success(Strategy, TM) is the arithmetic mean calculated after the strategy named 'Strategy' has lived one life in the world defined by the machine named 'TM'. The sum total is across all machines with 1000 states the worlds of which are interesting.

The above Global IQ practically cannot be calculated and it is the theoretical value which we try to approximate by means of the Local IQ.

The Local IQ therefore is:

$$\text{Local IQ}(Strategy) = \sum_{i=1}^{1000} \text{Success}(Strategy, TM_i)$$

Where $TM_i$ is $i$-th preselected test world (Turing machine).

## The final definition

**Definition:** An AI is any strategy the Local IQ of which exceeds 0.7.

Here we selected the same value as we did in [3]. Both here and in [3] the value (0.7) is selected arbitrarily. Likewise, a corporation may announce that it is looking for a CEO and will be happy to employ anyone who solves 70% of the problems given in the test. The corporation may have to adjust this level later if the bar proves too low or too high.

**Definition:** An AI is any computer program if the strategy it plays in the first 1000 games is an AI strategy.

## Analogy with the Grandmaster definition

We will provide another explanation of the AI definition by repeating the same construct, but this time we will define a chess playing program. We have already done so in [4], but as the present study revisits and updates the construct described in [3], here we will also revisit the construct in [4] and will reflect the relevant adjustments.

We will define a chess playing program as a program which plays like a grandmaster. If someone wants to be a grandmaster, he or she should earn an ELO rating (calculated by the ELO rating system) of at least 2500. The bad news is that the ELO rating calculation is based on his or her performance in playing against other people. To obtain an objective assessment of how good is the player, we will replace the other players with a finite set of deterministic computer programs. Then our chess player will play one game against each of these other players and his or her result will be the arithmetic mean of the results achieved in the individual games. In case a player halts (goes cycling) and does not make it until the end of game we will award a utility victory to the opponent. The same we will do if the player plays incorrect move. That is, a program which aspires to become a grandmaster must play correct moves and had better avoid cycling because in the opposite case we will penalize it with a utility loss. This applies both to the chess playing program and to computer programs in the finite set against which we assess the performance of our program.

Which will be the finite set of deterministic computer programs that used in the grandmaster test? We can take all programs that do not exceed a predefined length. Most of these will play randomly, will go cycling often and will make many incorrect moves. It would be reasonable to screen them and leave only the interesting ones (programs that play too mindlessly are a ballast which only makes the test more cumbersome). Interesting programs will be those that do not cycle, do not make incorrect moves and, on top of all, play relatively well.

When looking for interesting worlds to use in the AI test we used the brute-force search method. This method will cannot be used here because it is very unlikely to randomly hit a program which does not cycle and makes only correct moves. Hence, instead the set of all programs which do not exceed a certain length, we will take a specific set of programs whereby all programs are interesting (do not cycle, do not make incorrect moves and play relatively well).

Let us take a particular program, which calculates the next five moves and distinguishes between three types of positions: winning (victory within the next five moves), losing (loss within the next five moves) and undetermined (all other positions). Which move will this program play? It will pick randomly one of the winning positions. In the absence of a winning position, it will chose randomly one of the undetermined positions. If an undetermined position is also lacking, the program will go ahead randomly with a losing position.

What we described above is a non-deterministic program. If we reckon how many deterministic strategies this program can implement, they are far too many (but still finitely many, because the length of the game is limited). Each of these strategies is calculated by infinitely many programs, but we shall assume that for each strategy we have selected one program which calculates this strategy.

As a result, we obtained a huge set of programs and may be tempted to say that our ELO rating will be calculated on the basis of the games with all these programs. Unfortunately these programs are far too many and we cannot calculate the ELO in this way because of combinatorial explosion. Instead, let us select 1000 of these programs and calculate the ELO after 1000 games (one game played with each program). We will select these programs randomly, but will select them only once and will always calculate the ELO on the basis of the same test programs.

How can we make a deterministic program out of a non-deterministic one? That is easy, instead of picking a random move we pick a pseudorandom move. Before starting the game, we will initialize the random number generator with the number one (1). Thus we obtain one deterministic program. We need 1000 deterministic programs. We will initialize with the numbers from 1 to 1000 and will obtain 1000 different deterministic programs (among these 1000 there might be some identical programs, in particular where the random numbers generator is not as good as it should be). These will be the programs from which we will derive the ELO rating.

Let we say that a program will become a grandmaster if the so-calculated ELO is more than 90%. This is an arbitrary value picked by us. It may turn out that the value should be higher. We may even have to adjust the test worlds and ask them to calculate maybe the next 10 moves rather than 5 moves.

Then we stumble upon the crammer issue again. It is perfectly possible to learn by heart how to beat some of these 1000 programs. Remember these are deterministic programs and if we beat a deterministic program once we can beat it as many times as we want by replaying the same game.

A fairly good analogy emerges between the grandmaster and AI definitions. In the first case we refer to an ELO rating and in the latter case to an IQ rating. In the first case we play chess against 1000 opponents and in the latter case we live 1000 lives in 1000 worlds. The difference is that in the chess scenario we play one game against each opponent and in the IQ scenario we play one thousand games in each life. This is so because in the first case the rules of the game are fixed and it is expected that a grandmaster program knows the rules and knows how to play (rather than learning how to play while it plays). In the second case, the AI does not know the rules of life and needs one thousand games in order to discern these rules and learn how to live successfully.

# Conclusion

In this study we described an exam (test) which enables us calculate the IQ of any program. More precisely, we specified the program which will conduct the test and tell us what the candidate's IQ is. Given the deep granularity of our specification, we can now relax and ask some student to do the coding for us, perhaps as a course project.

This program will enable us do the AI test in the matter of minutes. This the time which the examiner program needs to calculate the test result. We should then add the thinking time we afford to each candidate. When we consider AI as a strategy, we need not ask the question of how much time the candidate will spend thinking. When we consider AI as a program which calculates an AI strategy, we must announce how much time we afford to the program for calculating one step.

So, the time for the test will be some minutes to calculate the test result, plus the thinking time which we afford to the candidate, plus some time for generation of the test (construction of the array of 1000 numbers). The last time we do not consider because it will be spend only once.

Let us think about what this test can be used for. If someone comes to us with a particular program, we would test it and say what the IQ of that program is. But, we do not have any AI-candidate programs. So we have nobody to apply this test to.

One possible application of the test described in this study is finding the AI. We might use the brute-force search method. Of course we can search with this method, but combinatorial explosion would not let us go too far with this method. There is, however, a smarter way of searching. We can construct a genetic algorithm. We will sit at one powerful computer, create a population of AI candidate programs, and calculate each candidate's IQ. We shall combine candidates with higher IQs in order to obtain offspring with even higher IQ. We shall kill the low-IQ candidates in order to make room for more promising programs. Using this natural selection approach we shall obtain programs with very high IQs.

The genetic algorithm is one way of finding the AI, but we will end up with a program the inner workings of which are enigmatic to us. If we are to control a program, we better write it ourselves rather than generate it automatically. For this reason, I am a proponent of the direct approach for creating AI. Therefore, I assert that we should write this program with our own hands.

# Acknowledgements

I wish to appreciate my colleagues Ivan Koychev, Anton Zinoviev and Andrey Sariev for the beneficial discussions on the AI Definition topic. Special acknowledgement to Ivan Koychev for his instrumental contribution to my summary of related work.

# References

[1] Dimiter Dobrev. First and oldest application. *1993.* http://dobrev.com/AI/first.html

[2] Dimiter Dobrev. AI - What is this. *PC Magazine - Bulgaria, November'2000, pp.12-13* (in Bulgarian, also in English at http://dobrev.com/AI/definition.html).

[3] Dimiter Dobrev. Formal Definition of Artificial Intelligence. *International Journal "Information Theories & Applications", vol.12, Number 3, 2005, pp.277-285.* http://www.dobrev.com/AI/

[4] Dimiter Dobrev. Parallel between definition of chess playing program and definition of AI. International Journal "Information Technologies & Knowledge ", vol.1, Number 2, 2007, pp.196-199.

[5] Dimiter Dobrev. Two fundamental problems connected with AI. *Proceedings of Knowledge - Dialogue - Solution 2007, June 18 - 25, Varna, Bulgaria, Volume 2, p.667.*

[6] Dimiter Dobrev. Giving the AI definition a form suitable for engineers. *April, 2013.* http://www.dobrev.com/AI/

[7] Alan Turing. Computing machinery and intelligence. *Mind, 1950.*

[8] John McCarthy. What is Artificial Intelligence? November, 2007. (www-formal.stanford.edu/jmc/whatisai/)

[9] Huazheng Wang, Fei Tian, Bin Gao, Jiang Bian, Tie-Yan Liu. Solving Verbal Comprehension Questions in IQ Test by Knowledge-Powered Word Embedding. *arXiv: 1505.07909, May, 2015.*

[10] Detterman, Douglas K. A Challenge to Watson. *Intelligence, v39 n2-3 p77-78 Mar-Apr 2011.*

[11] Feng Liu, Yong Shi, Ying Liu. Intelligence Quotient and Intelligence Grade of Artificial Intelligence. *arXiv: 1709.10242, September, 2017.*

[12] Dowe, D.L., & Hernández-Orallo, J. IQ tests are not for machines, yet. *Intelligence (2012), doi:10.1016/j.intell.2011.12.001*

[13] Hernández-Orallo, J., & Minaya-Collado, N. (1998). A formal definition of intelligence based on an intensional variant of Kolmogorov complexity.
Proc. intl symposium of engineering of intelligent systems (EIS'98), February
1998, La Laguna, Spain (pp. 146–163). : ICSC Press.

[14] Insa-Cabrera, J., Dowe, D. L., & Hernandez-Orallo, J. (2011). Evaluating a reinforcement learning algorithm with a general intelligence test. *In J. M. J. A. Lozano, & J. A. Gamez (Eds.), Current topics in artificial intelligence. CAEPIA 2011. : Springer (LNAI Series 7023).*